\pdfoutput=1
\documentclass[11pt]{article}

\usepackage{acl}

\usepackage{times}
\usepackage{latexsym}

\usepackage[T1]{fontenc}
\usepackage[utf8]{inputenc}

\usepackage{microtype}

\usepackage{inconsolata}

\usepackage{graphicx}

\usepackage{xspace}
\usepackage{adjustbox}
\usepackage{pgfplots} 
\usepackage{tcolorbox}
\usepackage[cachedir=minted-cache]{minted}
\usepackage{amsmath}
\usepackage{listings}
\usepackage{booktabs}
\usepackage{xcolor}
\usepackage{bm}
\usepackage{enumitem}

\usepackage{color}
\definecolor{keywordcolor}{rgb}{0.7, 0.1, 0.1}   % red
\definecolor{tacticcolor}{rgb}{0.0, 0.1, 0.6}    % blue
\definecolor{commentcolor}{rgb}{0.4, 0.4, 0.4}   % grey
\definecolor{symbolcolor}{rgb}{0.0, 0.1, 0.6}    % blue
\definecolor{sortcolor}{rgb}{0.1, 0.5, 0.1}      % green
\definecolor{attributecolor}{rgb}{0.7, 0.1, 0.1} % red
\definecolor{lightgray}{rgb}{0.95, 0.95, 0.95}

\usepackage[utf8]{inputenc}
\DeclareUnicodeCharacter{2115}{\ensuremath{\mathbb{N}}}
\usepackage{newunicodechar}
\newcommand{\method}{\textbf{{NFL-HR}}\xspace}

\title{Let's Reason Formally: Natural-Formal Hybrid Reasoning Enhances LLM's Math Capability}

\author{
 \textbf{Ruida Wang\textsuperscript{1}}\thanks{First authors},
 \textbf{Yuxin Li\textsuperscript{2}}\footnotemark[1],
 \textbf{Yi R. (May) Fung\textsuperscript{2}}, 
 \textbf{Tong Zhang\textsuperscript{1}} 
\\
\small{
    \textsuperscript{1}University of Illinois Urbana-Champaign, 
    \textsuperscript{2}Hong Kong University of Science and Technology 
}
\\
 \small{
    \textbf{Correspondence:} 
    \href{ruidaw@illinois.edu}{ruidaw@illinois.edu}, 
    \href{ylinq@connect.ust.hk}{ylinq@connect.ust.hk},
 } 
\\
 \small{
    \href{yrfung@ust.hk}{yrfung@ust.hk}, 
    \href{tongzhang@tongzhang-ml.org}{tongzhang@tongzhang-ml.org}
 }
}

\begin{document}
\maketitle

\maketitle
\begin{abstract}
    Enhancing the mathematical reasoning capabilities of LLMs has garnered significant attention in both the mathematical and computer science communities. Recent works have made substantial progress in both Natural Language (NL) reasoning and Formal Language (FL) reasoning by leveraging the potential of pure Reinforcement Learning (RL) methods on base models. However, RL approaches struggle to impart new capabilities not presented in the base model~\cite{yue2025does}, highlighting the need to integrate more knowledge like FL into NL math reasoning effectively. Yet, this integration is challenging due to inherent disparities in problem structure and reasoning format between NL and FL~\cite{wang2024theoremllama}. To address these challenges, we introduce \textbf{NL-FL HybridReasoning (NFL-HR)}, an end-to-end framework designed to incorporate the FL expert into NL math problem-solving. To bridge the NL and FL input format gap, we propose the \textit{NL-FL Problem Alignment} method, which reformulates the Question-Answering (QA) problems in NL as existence theorems in FL. Subsequently, the \textit{Mixed Problem Input} technique we provide enables the FL reasoner to handle both QA and existence problems concurrently. Lastly, we mitigate the NL and FL output format gap in reasoning through an LLM-based \textit{Answer Extraction} mechanism. Comprehensive experiments demonstrate that the \textbf{NFL-HR} framework achieves \textbf{89.80\%} and \textbf{84.34\%} accuracy rates on the MATH-500 and the AMC benchmarks, surpassing the NL baseline by 4.60\% and 4.82\%, respectively. Notably, some problems resolved by our framework remain unsolved by the NL baseline model even under a larger number of trials.

\end{abstract}

\section{Introduction}\label{sec:intro}

The capability of performing rigorous mathematical reasoning has always been regarded as a cornerstone of human intelligence and a fundamental goal of machine learning systems~\cite{newell1956logic}. Among these tasks, mathematical reasoning is considered crucial for evaluating the capabilities of Large Language Models (LLMs). Recently, both academia and industry have been actively working on two branches of LLM math reasoning, namely Natural Language (NL) reasoning and Formal Language (FL) reasoning.  

In the NL math reasoning, taking advantage of the vast NL data during pre-training, researchers have been continuously making progress in the field during the last few years. 
Many works tried to enhance LLMs' Math capability from various perspective~\cite{hendrycks2021measuring, numina_math_datasets, rafailov2023direct, wei2022chain}. Recent progress in pure Reinforcement Learning (RL) training on base models has notable success in solving IMO-level problems through reflection capability in Long Chain-of-Thought (CoT)~\cite{guo2025deepseek}. 
However, the latest works have shown that while RL methods fail to introduce novel capabilities beyond the base model~\cite{gandhi2025cognitive}. Consequently, integrating knowledge from other domains, such as FL reasoning, has emerged as a promising direction to enhance LLM's mathematical reasoning abilities.

\begin{figure*}[t]
    \vspace{-0.1in}
    \centering
    \includegraphics[width=0.99\linewidth]{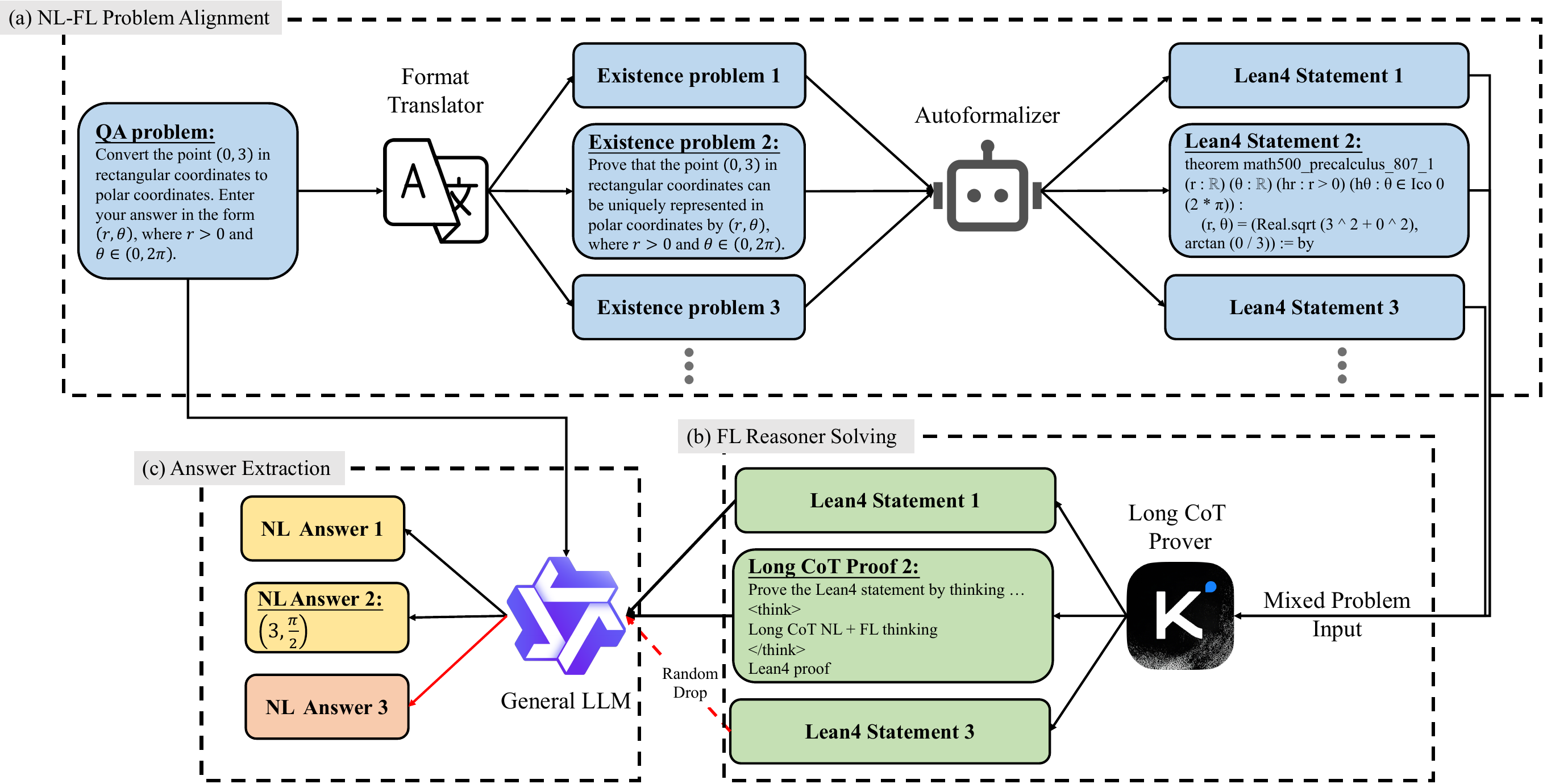}
    \caption{
        \textbf{NL-FL HybridReasoning (NFL-HR)} framework: 
        (a) NL-FL Problem Alignment: We first translate the QA-style NL problem into the NL existence problem using a general LLM, followed by converting the problem into an FL existence theorem through an autoformalizer.
        (b) FL Reasoner Problem Solving: We then apply the mixed problem input technique to ask the FL reasoner to concurrently address the QA problem in NL and the existence problem in FL within a unified Long CoT thinking process.
        (c) Answer Extraction: Finally, we use the LLM to extract the implicit NL answer from the FL prover's Long CoT output.
    }
    \label{fig:main}
    \vspace{-0.1in}
\end{figure*}

On the other end, FL reasoning leverages the inherent verifiability of formal systems like Lean~\cite{de2015lean, moura2021lean}, Coq~\cite{coq1996coq}, and Isabelle~\cite{paulson1994isabelle}. They provide a rigorous framework for defining solutions and verifying their correctness. 
This reliability facilitates large-scale and risk-free data generation~\cite{lin2025goedel, wang2025ma, xin2024deepseek, dong2025beyond}, which enables the use of synthetically generated data for stable RL/SFT training~\cite{wang2024theoremllama, xin2024deepseek1, wu2024internlm2}. 
With the advancements in Long CoT reasoning models, pure RL methods have proven to be particularly effective for FL reasoners~\cite{ren2025deepseek, wang2025kimina}. However, despite rapid progress in enhancing FL reasoning capabilities, most efforts remain confined to FL theorem-proving tasks, limiting the broader application of FL-derived knowledge to NL math problems.
A substantial gap persists between NL and FL reasoning due to the differences in problem structure and reasoning format. FL tasks are typically structured as closed-ended theorems with fixed conditions and goals. Whereas NL problems are often Question-Answering (QA) style, as those in MATH-500~\cite{lightman2023let}. Furthermore, FL training frequently biases models' final outputs toward FL code, reducing their effectiveness in directly addressing NL questions, which are often expressed in more concise and less structured formats.

In summary, in the context of NL math reasoning requiring broader knowledge integration to extend LLM capabilities, it is a natural idea to integrate FL reasoning to enhance that. However, FL reasoning remains constrained by its rigid problem structure and reasoning format. Addressing these challenges collectively presents a promising research direction in leveraging FL to enhance NL math problem-solving. 

To solve the problems above, we propose \textbf{NL-FL HybridReasoning (NFL-HR)}, an end-to-end framework that augments LLMs' NL math reasoning by enabling them to tackle problems that are otherwise difficult under a purely NL context. The framework overview is presented in Fig.~\ref{fig:main}. The core methodology of our framework can be summarized into three stages that works synergistically together:

\begin{enumerate}[leftmargin=13.6pt,topsep=2pt]
    \setlength\itemsep{0em}
    \item \textbf{NL-FL Problem Alignment:} We develop a method to establish a non-ambiguous alignment between the NL problem and FL problem. It is done by firstly converting QA problems into NL existence problems\footnote{The QA problems seek a specific answer while the existence problems ask for the proof of existence of the answer.} using a general LLM, followed by formalizing the existence problem into FL through an autoformalizer. 
    \item \textbf{FL Reasoner Solving:} At this stage, we implement the \textit{Mixed Problem Input} technique, allowing the FL reasoner to concurrently address both the FL existence problem and the QA problem in NL in its Long CoT thinking, thereby applying FL knowledge to solve NL tasks.
    \item \textbf{Answer extraction:} The final stage involves extracting the implicit NL answer from the Long CoT output generated by the FL reasoner. This step effectively unifying the output format of FL-derived reasoning with NL question-answering.
\end{enumerate}

We summarize our contributions as follows: (1) We introduce \method, a comprehensive framework that effectively incorporates FL reasoning to address NL math problems. (2) The proposed \textit{NL-FL Problem Alignment} and \textit{Answer Extraction} methods relatively effectively reduce the disparity between NL and FL reasoning in both problem structure and reasoning format, broadening the capability of FL methods. (3) We conduct comprehensive experiments using the \method framework, achieving accuracy rates of 89.80\% on the MATH-500 dataset and 84.34\% on the AMC dataset, surpassing the NL baseline of Qwen3-8B by \textbf{4.60\%} and \textbf{4.82\%}, respectively. 
Notably, the problems successfully solved by our framework are hard for the NL reasoner even with a larger number of trials, highlighting the effectiveness of integrating FL reasoning with NL problem-solving. Our code is available here \href{https://github.com/RickySkywalker/HybridReasoning}{here}.

\section{Methodology}\label{sec:meth}
In this section, we present the details of \textbf{NL-FL HybridReasoning} framework, which leverages the knowledge of Formal Language (FL) reasoners to tackle Natural Language (NL) math problems. To bridge the format gap between NL and FL reasoner's inputs, we propose the \textit{NL-FL Problem Alignment} method in Section~\ref{meth:align}.
We further introduce the \textit{Mixed Problem Input} technique in Section~\ref{meth:FL}, which enables the Long CoT FL reasoner to solve QA problem in NL and existence problem in FL concurrently.
Finally, we present the \textit{Answer Extraction} method to output format of FL reasoner to answer NL question in Section~\ref{meth:ans}.  Overall, the task for our framework can be defined as: ``Using post-training methods to extend the LLM's capabilities in solving NL-based math QA problems.''

\subsection{NL-FL Problem Alignment}~\label{meth:align} 
\vspace{-0.3in}
\subsubsection{QA-existence problem translation}~\label{align:trans}
A key challenge in applying FL reasoners to NL problems lies in the disparity between their problem structures. NL math datasets primarily contain blank-filling questions that require the LLMs to determine specific numerical or formulaic answers. For instance, a typical NL problem is structured as ``Given conditions $\{c_1, c_2, \cdots\}$, find the answer $a$,” as illustrated in Fig.~\ref{fig:QA_prob}, where the objective is to identify the value of $C$.

\begin{figure}[htbp]
\centering
\begin{adjustbox}{max width=0.99\linewidth}
\begin{tcolorbox}
Find the smallest positive real number $C$ for which
\[\left\| \begin{bmatrix} 2 & 3 \\ 0 & -2 \end{bmatrix}\textbf{v} \right\| \le C \|\textbf{v}\|\]
for all two-dimensional vectors $\textbf{v}.$
\end{tcolorbox}
\end{adjustbox}
\caption{QA problem for NL math dataset}\label{fig:QA_prob}
\end{figure}

On the other hand, in FL reasoning, the problem type is typically a closed-ended, proof-oriented task framed as: ``Given conditions $\{c_1, c_2, \cdots\}$, prove the goal $a$''. In NL-FL translation (or formalization), traditional methods demonstrated in \citet{zheng2021minif2f} will simply present the correct answer in the statement and let the FL reasoner prove the answer is correct. For instance, in the previous example, traditional methods will reformulate the problem as ``prove the $C$ is 4.'' While this is a valid formalization, it offers limited utility in deriving the answer we do not know.

However, if we directly use the LLM to transform the QA problem in NL into an FL problem, there is an ambiguous point. Since the FL requires a goal proof, but there is no such goal provided in the NL question. Thus, the LLMs tend to directly guess and answer to bridge the gap between the FL proof and the NL question. But this can confuse the FL reasoner since the guessed answer is mostly incorrect.

\begin{figure}[htbp]
\centering
\begin{adjustbox}{max width=0.99\linewidth}
\begin{tcolorbox}
\textbf{Prove that there exists} a smallest positive real number $C$ such that
$$\left\| \begin{bmatrix} 2 & 3 \\ 0 & -2 \end{bmatrix} \textbf{v} \right\| \le C \|\textbf{v}\|$$
for all two-dimensional vectors $\textbf{v}.$
\end{tcolorbox}
\end{adjustbox}
\caption{QA reformatted to existence problem}\label{fig:NL_existance}
\end{figure}

To address this, we adopt a few-shot prompting strategy to let a general LLM reformulate the QA problem into an NL existence problem. Instead of providing a specific value, this type of problem focuses on establishing the existence of the solution, thus removing the ambiguity. The detailed prompt is provided in Appendix~\ref{prompt:trans}, where we offer cross-domain examples to guide the transformation process. The reformulated NL existence problem for the QA task is illustrated in Fig.~\ref{fig:NL_existance}, where the objective shifts from ``finding the number'' to ``proving the existence of the number''.

\subsubsection{Autoformalization}~\label{align:auto}
In the FL research, many works try to automate the formalization process. They use millions of pairs NL-FL aligned problems to train LLMs named autoformalizer to do such a process~\cite{wang2025kimina, ren2025deepseek, lin2025goedel}. The autoformalizer's input is the NL proof problem and outputs its corresponding FL proof problem (or statement). The existence problem we derived above is the unambiguous counterparty of the QA problem in NL and is suitable for FL statement formulation. Thus, we can use the autoformalizer to transform the NL existence problem to the corresponding FL statement. Specifically, we adopt Lean4 in the \method framework due to its rapidly advancing models in autoformalization. An example of a formalized FL statement is demonstrated in Fig.~\ref{fig:FL_stat}.

\begin{figure}[htbp]
\centering
\begin{adjustbox}{max width=0.99\linewidth}
\begin{tcolorbox}
\begin{lstlisting}
theorem math500_precalculus_675_2 :
    ∃ (C : \R), 
    IsLeast {C | 0 < C ∧ ∀ (v : EuclideanSpace \R (Fin 2)), ‖(2 * v 0 + 3 * v 1, -2 * v 0)‖ ≤ C * ‖v‖} C := by sorry
\end{lstlisting}
\end{tcolorbox}
\end{adjustbox}
\caption{Formalized Lean4 statement}\label{fig:FL_stat}
\end{figure}

The autoformalization process effectively mitigates the format gap between NL and FL problems, enabling the FL reasoner to indirectly address the QA problem in NL systematically.

\subsection{FL Reasoner Problem Solving}~\label{meth:FL}
In this section, we introduce the \textit{Mixed Problem Input} technique, which enables the FL reasoner (or prover) to implicitly address the QA problem in NL while processing their corresponding FL existence problems. 
Formally, the input to the FL prover is defined as:
\[Prover(\bm{x}_{FL}, \bm{x}_{NL}) = \bm{z}_{CoT}, \bm{y}_{FL}\]
where $\bm{x}_{FL}$ and $\bm{x}_{NL}$ represent the FL and NL problems, $\bm{z}_{CoT}$ denotes the Long CoT content and $\bm{y}_{FL}$ is the FL solution. During the FL prover's reasoning process, we observed that it initially resolves the NL question included in the prompt before generating the corresponding FL proof. Notably, even when the NL problem mismatches the corresponding FL statement, the prover still tends to address the NL problem first. Further analysis of this behavior is provided in Appendix~\ref{appendix:Mixed_input}.

Thus, we take advantage of such behavior in the \textit{Mixed Problem Input} technique, where we use the QA problem as $\bm{x}_{NL}$ and the translated existence Lean4 statement as $\bm{x}_{FL}$. During the reasoning process, the FL prover initially addresses the QA problem, which indirectly provides an answer through its Long CoT. This response is then utilized to write Lean4 code to prove the FL existence problem. By structuring the input in a QA-existence mixed problem format, the FL prover leverages its expertise to resolve the NL question while concurrently generating the FL proof. An example of the input-output structure is presented in Fig.~\ref{fig:FL_reasoner_input_output}, and a detailed prompt can be found in Appendix~\ref{prompt:prover}.

\begin{figure}[htbp]
\centering
\begin{adjustbox}{max width=0.99\linewidth}
\begin{tcolorbox}
\begin{Verbatim}[breaklines=True]
== Input ==
Think about and solve the following problem step by step in Lean 4. 
# Problem: <NL_QA_problem>
# Formal statement
```lean4
/-- <NL_QA_problem> -/
<FL_existence_problem> 
== Output ==
<think>
# Finding the Smallest Positive Real Number $C$ for a Matrix Norm Inequality
<reasoning contents omitted>
**The answer is: 4**
# Now translated it to Lean4:
<reasoning contents omitted>
</think>
<FL solutions>
\end{Verbatim}
\end{tcolorbox}
\end{adjustbox}
\caption{Input-Output of the FL reasoner}\label{fig:FL_reasoner_input_output}
\end{figure}

However, in FL prover's reasoning process of the mixed input problem, the NL answer is only implicitly presented in the Long CoT and not correctly formatted. 
This is because the training process of the prover established a rigid FL format bias.
It makes the prover unable to directly provide the NL answer in a formatted way for verification.
Thus, we employ the answer extraction method to retrieve the desired response from the Long CoT content.

\subsection{Answer Extraction}~\label{meth:ans}
\vspace{-0.2em}
This section details the approach for retrieving the implicit solutions embedded within the Long CoT content and enabling the framework to combine the knowledge from the NL and FL reasoners. Specifically, we use a general-purpose LLM to extract the NL answer from the Long CoT output, formatting the response in the \textbackslash{boxed}\{\} block for verification. The prompt structure is illustrated in Fig.~\ref{fig:prompt_answer_ext}, with the complete prompt template provided in Appendix~\ref{prompt:ans}.

\begin{figure}[htbp]
\centering
%\small
\begin{adjustbox}{max width=0.99\linewidth}
\begin{tcolorbox}
\begin{Verbatim}[breaklines=True]
Find the answer to the following question in the provided long CoT content. Your answer should be in \boxed\{\} format.
Here is the question:
<NL_QA_problem>
The answer is contained in the following Long CoT content:
<Long_CoT_from_prover>
\end{Verbatim}
\end{tcolorbox}
\end{adjustbox}
\caption{Prompt for answer extraction }\label{fig:prompt_answer_ext}
\vspace{-0.2em}
\end{figure}
To prevent overthinking in the straightforward answer extraction task, we do not use the thinking mode of the general LLM. But we do allow the LLM to perform the normal CoT to check the correctness of the answer. Additionally, to reduce potential biases introduced during FL training, we randomly exclude some FL outputs and prompt the NL model to directly resolve the problem, allowing the framework to leverage both the general LLM and the FL reasoner's knowledge
\section{Experiments}\label{sec:exp}
\vspace{-0.2em}
To validate the effectiveness of our proposed \textbf{NL-FL HybridReasoning} framework, we conduct comprehensive experiments on widely applied benchmarks~\cite{lightman2023let, numina_math_7b}. Additionally, we evaluate the framework's unique FL capabilities (Section~\ref{exp:unique}), the computation cost of \method (Section~\ref{exp:comp}), and conduct ablation studies (Section~\ref{exp:abl}) to further substantiate its effectiveness. 

\begin{table*}
    \vspace{-0.15in}
    \centering
    \resizebox{\textwidth}{!}{
    \tiny
    \begin{tabular}{ccccc}
        \hline
        \rule{0pt}{1.06\normalbaselineskip}
        \textbf{\scalebox{0.95}{Dataset}}    & \textbf{\scalebox{0.95}{Data Number}}  & \textbf{\scalebox{0.95}{NL Reasoning}}    & \textbf{\scalebox{0.95}{NL-FL HybridReasoning}} & \textbf{\scalebox{0.95}{Improvement}}  \\[0.2em]
        \hline
        \rule{0pt}{1.03\normalbaselineskip}
        \textbf{\scalebox{0.95}{AMC}    }    & \scalebox{0.95}{83}                    & \scalebox{0.95}{79.52\%}                  & \textbf{\scalebox{0.95}{84.34\%}}         & \scalebox{0.95}{4.82\%} \\[0.2em]
        \hline
        \rule{0pt}{1.07\normalbaselineskip}\textbf{\scalebox{0.95}{MATH-500}}   & \scalebox{0.95}{500}                   & \scalebox{0.95}{85.20\%}                   & \scalebox{0.95}{\textbf{89.80\%}}        & \scalebox{0.95}{4.60\%} \\[0.2em]
        \hline
        \rule{0pt}{1.06\normalbaselineskip}\textit{\scalebox{0.9}{MATH-500 by subject}} \\
        \textbf{\scalebox{0.95}{Prealgebra}} & \scalebox{0.95}{82}                   & \scalebox{0.95}{84.15\%}                 & \textbf{\scalebox{0.95}{89.02\%}}          & \scalebox{0.95}{4.88\%} \\
        \textbf{\scalebox{0.95}{Counting \& Probability}}    & \scalebox{0.95}{38}    & \scalebox{0.95}{84.21\%}                 & \textbf{\scalebox{0.95}{89.47\%}} & \scalebox{0.95}{5.26\%} \\
        \textbf{\scalebox{0.95}{Intermediate Algebra}}  & \scalebox{0.95}{97}       & \scalebox{0.95}{84.54\%}                 & \textbf{\scalebox{0.95}{89.69\%}}          & \scalebox{0.95}{5.15\%} \\
        \textbf{\scalebox{0.95}{Geometry}}   & \scalebox{0.95}{41}                   & \scalebox{0.95}{65.85\%}                  & \textbf{\scalebox{0.95}{80.49\%}}       & \scalebox{0.95}{14.63\%} \\
        \textbf{\scalebox{0.95}{Precalculus}}    & \scalebox{0.95}{56}                & \scalebox{0.95}{76.79\%}                  & \textbf{\scalebox{0.95}{83.93\%}}        & \scalebox{0.95}{7.14\%} \\
        \textbf{\scalebox{0.95}{Algebra}}    & \scalebox{0.95}{124}                  & \scalebox{0.95}{94.35\%}                  & \textbf{\scalebox{0.95}{95.16\%}}       & \scalebox{0.95}{0.81\%} \\
        \textbf{\scalebox{0.95}{Number Theory}}  & \scalebox{0.95}{62}                & \scalebox{0.95}{90.32\%}                  & \textbf{\scalebox{0.95}{91.94\%}}          & \scalebox{0.95}{1.61\%} \\[0.2em]
        \hline
    \end{tabular}
    }
    \vspace{-0.02in}
    \caption{Main experiment result of \method under MATH-500 and AMC benchmarks with pass@16 metric.}
    \label{tab:main}
    \vspace{-0.15in}
\end{table*}

\subsection{Experiment Setup}\label{exp:setup}

\subsubsection{Dataset}\label{setup:data}
\vspace{-0.04em}
To evaluate the efficiency of \method in advancing LLM NL math reasoning, we choose the MATH-500~\cite{lightman2023let} and AMC~\cite{numina_math_7b} datasets.

MATH-500 is a collection of 500 challenging mathematical problems designed to benchmark advanced reasoning and problem-solving capabilities in LLMs. The dataset covers diverse mathematical domains, with each problem accompanied by its corresponding numerical or formulaic solution for verification. It is widely used for research in mathematical reasoning, symbolic computation, and automated theorem proving. The AMC dataset consists of 83 problems extracted and reformatted from AMC-2022 and AMC-2023~\cite{numina_math_7b}, with a format transition from multiple-choice to QA-style numerical answers. Positioned as an intermediate-level dataset between MATH-500 and IMO-level problems, making it a suitable benchmark for assessing more advanced reasoning capabilities in NL-based math tasks.

\subsubsection{Baseline \& Evaluation  Metric}\label{setup:base}
We choose Qwen3-8B~\cite{qwen3} as the baseline model due to its advanced Long CoT reasoning capability and leading performance across various benchmarks. We do not apply RL methods to the NL reasoning model because Long CoT models are typically distilled from RL-trained large models~\cite{qwen3, guo2025deepseek}, as shown by~\citet{guo2025deepseek}; RL methods applied directly to small models typically do not outperform the distilled variants.

For evaluation, we utilize the pass@16 metric on both datasets. This metric asks the model to generate 16 potential answers for each question and considers the problem correctly solved if any of the generated answers match the correct solution. It is an appropriate metric to evaluate the capability limit for math reasoning frameworks. We chose this metric to align with the recent findings that pure RL methods reduce incorrect outputs but struggle to expand overall reasoning capability~\cite{yue2025does, gandhi2025cognitive}, which makes the assessment of capability limit meaningful.

\subsection{Implementation Details}\label{exp:imp}
\vspace{-0.2em}
We employ Qwen3-8B~\cite{qwen3} with Long CoT thinking mode for format translation and non-thinking mode for answer extraction. For autoformalization, we utilize Kimina-Autoformalizer~\cite{wang2025kimina}, while Kimina-Prover-Preview-7B~\cite{wang2025kimina} serves as the FL reasoner. Details of the Kimina-Prover series of models can be found in Appendix~\ref{appendix:detail_models}. The temperature for proof generation is configured at 0.7, with other inference steps set to 0.6 to maintain a balance between exploration and stability. We also set the temperature of the baseline model to 0.7. Inference is conducted using vllm~\cite{kwon2023efficient} with default parameters. To manage GPU usage, we limit the maximum token generation to 8192 tokens per query. The complete evaluation process consumes approximately 200 A6000 GPU hours.

\subsection{Main Results}\label{exp:result}
\vspace{-0.3em}
We present the main experiment results in Tab.~\ref{tab:main}. From the table, the \method achieves accuracy rates of 84.34\% on the AMC dataset and 89.80\% on the MATH-500 dataset. Outperforming the NL baseline by 4.82\% and 4.60\% respectively, which records 79.52\% on AMC and 85.20\% on MATH-500.

In the subjective-level result, the NL reasoning baseline performs relatively badly on domains like Geometry and Precalculus.
In Precalculus, the NL model achieves only 76.79\%, while \method reaches 83.93\%. This gain stems from FL reasoning's emphasis on the precise, step-by-step logical analysis, which is vital for handling complex calculations in calculus.
In Geometry, since most problems are about numerical geometrical concepts like degrees and edges, whose data is easy to obtain by FL training in analytic geometry, but not that easy through NL training. This advantage leads to a 14.63\% of improvement.

From a general perspective, the NL baseline exhibits substantial performance variability across subjects. It excels in areas with abundant data, such as algebra and number theory, but falters in fields that require precise logical deductions, like geometry and precalculus.
In contrast, \method demonstrates consistent and relatively high performance across all domains. It effectively mitigates the weaknesses of NL reasoning by incorporating the rigorous verification capabilities of FL during training.
This consistent improvement underscores the value of incorporating FL-derived logical structures into NL problem-solving.

\subsection{Study on FL's Unique Capability}\label{exp:unique}
\begin{table}
    \centering
    \resizebox{\linewidth}{!}
    {\begin{tabular}{ccccc}
    \toprule
         \textbf{Dataset}   & \textbf{Subset problem \#}    & \textbf{NL-16}     & \textbf{NL-64}    & \textbf{HR-16} \\
         \midrule
         \textbf{AMC}       & 7                             & 0\%                & 0\%               & 100\% \\ 
         \textbf{MATH}      & 23                            & 0\%                & 0\%               & 100\% \\ 
    \bottomrule
    \end{tabular}}
    \vspace{-0.05in}
    \caption{Results on FL's unique capability, where the ``Subset problem \#'' means the number of problems not solved by the NL reasoning but solved by \method (HR) in the main experiment.}
    \label{tab:FL_unique}
    \vspace{-0.1in}
\end{table}
To verify the FL prover supplies abilities hard to attain through pure NL training, we re-evaluated the NL baseline on the subset of problems that are only solved by \method in the main experiment. This subset represents 8.43\% of AMC problems and 4.60\% of MATH-500 problems. To test whether these problems can be finished by NL reasoner, we allow a \textit{pass@64} generation in this experiment. The results are reported in Tab.~\ref{tab:FL_unique}.

The table shows that even with 64 attempts, the NL models obtained a 0\% accuracy rate on every problem for both sub-datasets. It typically stalls in the self-reflection loop and produces no valid answer. In contrast, \method continues to solve all items in the group in the pass@16 metric, confirming that the FL component contributes distinctive knowledge that the NL system alone finds hard to cover.

\subsection{Computation Cost Study}\label{exp:comp}
We test the efficiency of the \method framework by comparing the average tokens generated by \method and the NL reasoning baseline of Qwen3-8B in the MATH-500 and AMC dataset under the pass@16 metric. The results are presented in Tab.~\ref{tab:comp}. We can see that the \method requires approximately 850 additional tokens per query on average. Compared to the unique capability obtained of \method as shown in Section~\ref{exp:unique}, where \method can solve problems that the NL baseline is unable to answer even under much larger trials. This demonstrates the gain of capability that outweighs the modest increase in token usage. 

Furthermore, in real-time inference, both systems cost the same scale of GPU hours on average. This is because the \method distributes its generation across multiple stages, but the NL baseline conducts a single-pass generation that has more GPU cost as it generates later tokens. Therefore, we conclude the additional generation cost as a reasonable and worthwhile trade-off for the improved reasoning performance.

\begin{table}
    \centering
    \small
    {\begin{tabular}{cccc}
    \toprule
         \textbf{Method}            & \textbf{AMC}      & \textbf{MATH-500}     & \textbf{Average}  \\
         \midrule
         \textbf{Qwen3-8B}          & 6,611.53          & 4,305.30              & 5,458.42 \\
         \textbf{\method}           & 7,447.06          & 5,173.63              & 6,310.35 \\
    \bottomrule
    \end{tabular}}
    \vspace{-0.05in}
    \caption{Tokens generated for Qwen3-8B and \method in AMC and MATH-500 dataset under pass@16.}
    \label{tab:comp}
\end{table}

\begin{table}
    \centering
    \resizebox{\linewidth}{!}
    {\begin{tabular}{ccc}
    \toprule
         \textbf{Method}                        & \textbf{AMC}      & \textbf{MATH-500} \\
         \midrule
         \textbf{NL Reasoning}                  & 79.52\%           & 85.20\%  \\
         \textbf{\method w/o existence align.}  & 77.11\%           & 88.20\% \\
         \textbf{\method w/o expert prover}     & 80.72\%           & 88.33\% \\
         \textbf{\method}                       & 84.34\%           & 89.80\% \\
    \bottomrule
    \end{tabular}}
    \vspace{-0.05in}
    \caption{Ablation study results, all experiments are performed under pass@16.}
    \label{tab:abl}
    \vspace{-0.25in}
\end{table}
\subsection{Ablation Studies}\label{exp:abl}

\subsubsection{Drop existence alignment}\label{abl:trans}

To test the effectiveness of the NL-FL Problem Alignment, we drop the existence alignment step in \method and replace it with directly passing the NL QA-style problems to the autoformalizer to translate. 
During this experiment, it was observed that such direct formalization could lead to discrepancies between the correct NL answer and the generated FL statement, potentially disorienting the prover.
Consequently, in this ablated configuration, the prover was guided only to solve the autoformalized statement, with only the FL reasoning segment subsequently utilized for answer extraction.
The dropping of the existence alignment causes the accuracy rate to reduce by 4.42\%.
Notably, on the harder AMC dataset, this performance deterioration was particularly acute, with the framework's accuracy diminishing to levels below that of pure NL reasoning. This suggests that increasingly sophisticated datasets require more rigorous reasoning capabilities. Thereby heightening the importance of precise FL integration facilitated by the alignment step.

\subsubsection{Drop expert FL prover}\label{abl:prover}

To evaluate the contribution of the specialized FL prover, this experiment involved substituting it with the general-purpose LLM (Qwen3-8B) in \method. The corresponding results are detailed in Tab.~\ref{tab:abl}. This configuration yielded an average performance decrease of 2.55\% compared to the original \method, underscoring the efficacy of incorporating an FL expert for enhancing NL reasoning. This performance dip is attributed to the general LLM's lack of field-specific FL training, which consequently impeded the framework's ability to address advanced problems in certain subjects. Nevertheless, even with this substitution, the framework demonstrated an average improvement of 2.17\% over the NL baseline. This suggests that activating the model's inherent FL knowledge can still benefit NL reasoning.

\subsection{Case Study}\label{exp:case}

This section presents a detailed case study illustrating each stage of the \method framework. Due to the space constraints, we only provide analysis in the main paper, while the detailed input-output examples are available in Appendix~\ref{appendix:in-iout}.

During the QA-existence problem translation stage, the LLM demonstrates a thorough analysis of the task, and the model also effectively utilizes the provided few-shot examples. Notably, even when presented with a QA problem in the sense of unit translation, a format not typically framed as an existence problem in NL math datasets. The model achieves an accurate translation. And provide the result in a markdown box as instructed. In the subsequent autoformalization stage, the autoformalizer successfully converts the NL existence problem into a formal Lean4 statement. Despite the inclusion of some unnecessary elements in the formal goal, it is still a valid autoformalization, and the result successfully compiles.

For the FL prover generation phase, the expert prover correctly and concisely addresses the NL question. It then generates the corresponding FL proof, which also passes verification.
Finally, in the answer extraction stage, even without Long CoT, the general LLM still performs chain-of-thought reasoning to verify the result based on the FL prover's output, as intended.
This behavior enhances the framework's robustness and can potentially rectify minor discrepancies arising from the FL reasoning process.
\section{Related Work}\label{sec:relat}

\subsection{Natural Math Reasoning}\label{relat:NL}

Enhancing the mathematical problem-solving abilities of LLMs has consistently attracted interest from both industrial and academic research communities. This is largely because mathematical reasoning is widely considered a hallmark of advanced intelligence. From their interception, LLMs demonstrated notable aptitude for mathematical tasks. Benchmark datasets have charted rapid advancements, progressing from primary-school level problems~\cite{cobbe2021gsm8k}, through high-school problems~\cite{lightman2023let, hendrycks2021measuring}, to university-competition-level questions~\cite{gulati2024putnam}. 
Concurrently, the development of extensive datasets such as OpenWebMath~\cite{paster2023openwebmath}, ProofPiles~\cite{azerbayev2023llemma}, and Numina-Math~\cite{numina_math_datasets} has expanded the scale of the math corpus to billions of tokens for pre-training and millions of question-answering pairs for fine-tuning. Alongside these data curation efforts, researchers have explored diverse training methodologies, ranging from simple instruction fine-tuning~\cite{brown2020language} and code-augmented training~\cite{roziere2023code} to Reinforcement Learning (RL) alignment strategies~\cite{rafailov2023direct, schulman2017proximal, shao2024deepseekmath}. More recently, pure RL training applied to base models to develop Long Chain-of-Thought capabilities has yielded impressive performance gains, exemplified by DeepSeek-R1~\cite{guo2025deepseek}. However, recent studies suggest that such RL-based training primarily refines existing abilities rather than imparting novel capabilities beyond those inherent in the base model~\cite{yue2025does, gandhi2025cognitive}. Consequently, investigating methods to integrate diverse knowledge sources and reasoning paradigms into LLM mathematical problem-solving has emerged as a significant research direction.

\subsection{Formal Math Reasoning}\label{relat:FL}

Formal Language (FL) reasoning involves expressing mathematical statements in first-order logic, rendering every component of the mathematical reasoning system verifiable. This approach mitigates ambiguity and provides a solid foundation for the reasoning process. Researchers have developed many FLs in the last decades, such as Isabelle~\cite{paulson1994isabelle}, Lean~\cite{de2015lean, moura2021lean}, CoQ~\cite{coq1996coq}, Metamath~\cite{megill2019metamath}, and HOL Light~\cite{harrison2009hol}. Among these, Lean4~\cite{moura2021lean} has garnered significant attention due to its elegance and simplicity.

Most research in FL reasoning has concentrated on developing more advanced prover models. Researchers have proposed various methods to enhance the formal reasoning capabilities of LLMs. Representative approaches include LeanDojo~\cite{yang2024leandojo}, which applies retrieval methods to suggest tactics; TheoremLlama~\cite{wang2024theoremllama}, which attempts to transform NL proofs into FL proofs; and the DeepSeek-Prover family~\cite{xin2024deepseek, xin2024deepseek1, ren2025deepseek}, Goedel-Prover~\cite{lin2025goedel}, and STP~\cite{dong2025beyond}, which employ advanced techniques for large-scale formal data annotation. Recently, systems like MA-LoT~\cite{wang2025ma}, Kimina-Prover~\cite{wang2025kimina}, and DeepSeek-Prover-V2~\cite{ren2025deepseek} utilize different strategies to train models with Long Chain-of-Thought, advancing formal reasoning capabilities to address IMO-level problems.

Despite these rapid advancements, inherent disparities between FL and NL in both problem formulation and reasoning paradigms mean that few studies have directly leveraged FL models to enhance the NL reasoning capabilities of LLMs. The majority of these efforts focus on answer selection based on formal verification~\cite{yao2025fans, zhou2024don}, or involve the indirect incorporation of FL-derived knowledge during RL training~\cite{guo2025deepseek}. Concurrently, new benchmarks are emerging to explore the integration of formal reasoning with even more complex input modalities, such as the multimodal problems presented in MATP-Bench~\cite{he2025matp}. Consequently, bridging the gap between NL and FL reasoning represents a promising direction in the background of the general need to integrate diverse knowledge sources into math reasoning.
\section{Conclusion}\label{sec:conc}

This paper introduces \method, an end-to-end framework that directly integrates the knowledge of Formal Language (FL) into Natural Language (NL) mathematical problem-solving. The framework largely addresses the gap in input format between NL and FL problems through the proposed \textit{NL-FL problem Alignment} method, which transforms the QA problem in NL into an existence theorem-proving statement in FL. Subsequently, we present the method to solve NL questions by the FL reasoner using the \textit{Mixed Problem Input} technique. Finally, an answer extraction mechanism retrieves solutions from the FL reasoning process, which bridges the output format gap between NL and FL reasoning.
Comprehensive experiments demonstrate the effectiveness of \method, which achieves 89.80\% accuracy rate on the MATH-500 dataset and 84.34\% on the AMC dataset. Furthermore, \method surpasses the NL baseline across all subjects in the MATH-500 dataset and demonstrates the ability to solve problems relatively hard for pure NL reasoners. The code will be open-sourced to encourage further research and development in this area.
\section{Discussion}

Although pure RL training provides LLMs with strong capabilities in many cases, recent works have proved that such training is unable to provide the model with the knowledge that is relatively weak in its base model. 
This highlights the promising direction of research into methods that incorporate broader knowledge into NL reasoning.
Generally speaking, \method provides a promising and general setting for integrating expert knowledge in a specific field into reasoning. Thus, the contribution of this paper is not limited to the field of math reasoning but gives a general pipeline for the integration of different fields' knowledge.

\newpage

\section*{Limitations}

Despite the promising results of \method, there are still some limitations in our work that can be further addressed in future research. First, even though our work integrates FL methods in NL reasoning, the verifiability of FL is not properly used in the framework. Future work can focus on methods to fix NL reasoning through FL feedback. 
Secondly, our framework effectively bridges the gap between NL and FL reasoners' input and output formats. The gap may still hinder performance,  which makes the development of blank-filling FL suitable for QA problems a promising direction. 
In the sense of potential risks, there is no immediately identifiable risk for the results of this paper.

\bibliography{latex/custom}

\begin{thebibliography}{44}
\providecommand{\natexlab}[1]{#1}

\bibitem[{Azerbayev et~al.(2023{\natexlab{a}})Azerbayev, Piotrowski, Schoelkopf, Ayers, Radev, and Avigad}]{azerbayev2023proofnet}
Zhangir Azerbayev, Bartosz Piotrowski, Hailey Schoelkopf, Edward~W Ayers, Dragomir Radev, and Jeremy Avigad. 2023{\natexlab{a}}.
\newblock Proofnet: Autoformalizing and formally proving undergraduate-level mathematics.
\newblock \emph{arXiv preprint arXiv:2302.12433}.

\bibitem[{Azerbayev et~al.(2023{\natexlab{b}})Azerbayev, Schoelkopf, Paster, Santos, McAleer, Jiang, Deng, Biderman, and Welleck}]{azerbayev2023llemma}
Zhangir Azerbayev, Hailey Schoelkopf, Keiran Paster, Marco~Dos Santos, Stephen McAleer, Albert~Q Jiang, Jia Deng, Stella Biderman, and Sean Welleck. 2023{\natexlab{b}}.
\newblock Llemma: An open language model for mathematics.
\newblock \emph{arXiv preprint arXiv:2310.10631}.

\bibitem[{Beeching et~al.(2024)Beeching, Huang, Jiang, Li, Lipkin, Qina, Rasul, Shen, Soletskyi, and Tunstall}]{numina_math_7b}
Edward Beeching, Shengyi~Costa Huang, Albert Jiang, Jia Li, Benjamin Lipkin, Zihan Qina, Kashif Rasul, Ziju Shen, Roman Soletskyi, and Lewis Tunstall. 2024.
\newblock Numinamath 7b tir.
\newblock \url{https://huggingface.co/AI-MO/NuminaMath-7B-TIR}.

\bibitem[{Brown et~al.(2020)Brown, Mann, Ryder, Subbiah, Kaplan, Dhariwal, Neelakantan, Shyam, Sastry, Askell et~al.}]{brown2020language}
Tom Brown, Benjamin Mann, Nick Ryder, Melanie Subbiah, Jared~D Kaplan, Prafulla Dhariwal, Arvind Neelakantan, Pranav Shyam, Girish Sastry, Amanda Askell, and 1 others. 2020.
\newblock Language models are few-shot learners.
\newblock \emph{Advances in neural information processing systems}, 33:1877--1901.

\bibitem[{Cobbe et~al.(2021)Cobbe, Kosaraju, Bavarian, Chen, Jun, Kaiser, Plappert, Tworek, Hilton, Nakano, Hesse, and Schulman}]{cobbe2021gsm8k}
Karl Cobbe, Vineet Kosaraju, Mohammad Bavarian, Mark Chen, Heewoo Jun, Lukasz Kaiser, Matthias Plappert, Jerry Tworek, Jacob Hilton, Reiichiro Nakano, Christopher Hesse, and John Schulman. 2021.
\newblock Training verifiers to solve math word problems.
\newblock \emph{arXiv preprint arXiv:2110.14168}.

\bibitem[{Coq(1996)}]{coq1996coq}
Projet Coq. 1996.
\newblock The coq proof assistant-reference manual.
\newblock \emph{INRIA Rocquencourt and ENS Lyon, version}, 5.

\bibitem[{De~Moura et~al.(2015)De~Moura, Kong, Avigad, Van~Doorn, and von Raumer}]{de2015lean}
Leonardo De~Moura, Soonho Kong, Jeremy Avigad, Floris Van~Doorn, and Jakob von Raumer. 2015.
\newblock The lean theorem prover (system description).
\newblock In \emph{Automated Deduction-CADE-25: 25th International Conference on Automated Deduction, Berlin, Germany, August 1-7, 2015, Proceedings 25}, pages 378--388. Springer.

\bibitem[{Dong and Ma(2025)}]{dong2025beyond}
Kefan Dong and Tengyu Ma. 2025.
\newblock Beyond limited data: Self-play llm theorem provers with iterative conjecturing and proving.
\newblock \emph{arXiv preprint arXiv:2502.00212}.

\bibitem[{Gandhi et~al.(2025)Gandhi, Chakravarthy, Singh, Lile, and Goodman}]{gandhi2025cognitive}
Kanishk Gandhi, Ayush Chakravarthy, Anikait Singh, Nathan Lile, and Noah~D Goodman. 2025.
\newblock Cognitive behaviors that enable self-improving reasoners, or, four habits of highly effective stars.
\newblock \emph{arXiv preprint arXiv:2503.01307}.

\bibitem[{Gulati et~al.(2024)Gulati, Miranda, Chen, Xia, Fronsdal, de~Moraes~Dumont, and Koyejo}]{gulati2024putnam}
Aryan Gulati, Brando Miranda, Eric Chen, Emily Xia, Kai Fronsdal, Bruno de~Moraes~Dumont, and Sanmi Koyejo. 2024.
\newblock Putnam-axiom: A functional and static benchmark for measuring higher level mathematical reasoning.
\newblock In \emph{The 4th Workshop on Mathematical Reasoning and AI at NeurIPS'24}.

\bibitem[{Guo et~al.(2025)Guo, Yang, Zhang, Song, Zhang, Xu, Zhu, Ma, Wang, Bi et~al.}]{guo2025deepseek}
Daya Guo, Dejian Yang, Haowei Zhang, Junxiao Song, Ruoyu Zhang, Runxin Xu, Qihao Zhu, Shirong Ma, Peiyi Wang, Xiao Bi, and 1 others. 2025.
\newblock Deepseek-r1: Incentivizing reasoning capability in llms via reinforcement learning.
\newblock \emph{arXiv preprint arXiv:2501.12948}.

\bibitem[{Harrison(2009)}]{harrison2009hol}
John Harrison. 2009.
\newblock Hol light: An overview.
\newblock In \emph{International Conference on Theorem Proving in Higher Order Logics}, pages 60--66. Springer.

\bibitem[{He et~al.(2025)He, Lyu, Chen, Guo, and Fung}]{he2025matp}
Zhitao He, Zongwei Lyu, Dazhong Chen, Dadi Guo, and Yi~R Fung. 2025.
\newblock Matp-bench: Can mllm be a good automated theorem prover for multimodal problems?
\newblock \emph{arXiv preprint arXiv:2506.06034}.

\bibitem[{Hendrycks et~al.(2021)Hendrycks, Burns, Kadavath, Arora, Basart, Tang, Song, and Steinhardt}]{hendrycks2021measuring}
Dan Hendrycks, Collin Burns, Saurav Kadavath, Akul Arora, Steven Basart, Eric Tang, Dawn Song, and Jacob Steinhardt. 2021.
\newblock Measuring mathematical problem solving with the math dataset.
\newblock \emph{arXiv preprint arXiv:2103.03874}.

\bibitem[{Kwon et~al.(2023)Kwon, Li, Zhuang, Sheng, Zheng, Yu, Gonzalez, Zhang, and Stoica}]{kwon2023efficient}
Woosuk Kwon, Zhuohan Li, Siyuan Zhuang, Ying Sheng, Lianmin Zheng, Cody~Hao Yu, Joseph~E. Gonzalez, Hao Zhang, and Ion Stoica. 2023.
\newblock Efficient memory management for large language model serving with pagedattention.
\newblock In \emph{Proceedings of the ACM SIGOPS 29th Symposium on Operating Systems Principles}.

\bibitem[{LI et~al.(2024)LI, Beeching, Tunstall, Lipkin, Soletskyi, Huang, Rasul, Yu, Jiang, Shen, Qin, Dong, Zhou, Fleureau, Lample, and Polu}]{numina_math_datasets}
Jia LI, Edward Beeching, Lewis Tunstall, Ben Lipkin, Roman Soletskyi, Shengyi~Costa Huang, Kashif Rasul, Longhui Yu, Albert Jiang, Ziju Shen, Zihan Qin, Bin Dong, Li~Zhou, Yann Fleureau, Guillaume Lample, and Stanislas Polu. 2024.
\newblock Numinamath.
\newblock \url{[https://huggingface.co/AI-MO/NuminaMath-CoT](https://github.com/project-numina/aimo-progress-prize/blob/main/report/numina_dataset.pdf)}.

\bibitem[{Lightman et~al.(2023)Lightman, Kosaraju, Burda, Edwards, Baker, Lee, Leike, Schulman, Sutskever, and Cobbe}]{lightman2023let}
Hunter Lightman, Vineet Kosaraju, Yuri Burda, Harrison Edwards, Bowen Baker, Teddy Lee, Jan Leike, John Schulman, Ilya Sutskever, and Karl Cobbe. 2023.
\newblock Let's verify step by step.
\newblock In \emph{The Twelfth International Conference on Learning Representations}.

\bibitem[{Lin et~al.(2025)Lin, Tang, Lyu, Wu, Lin, Yang, Li, Xia, Chen, Arora, and Jin}]{lin2025goedel}
Yong Lin, Shange Tang, Bohan Lyu, Jiayun Wu, Hongzhou Lin, Kaiyu Yang, Jia Li, Mengzhou Xia, Danqi Chen, Sanjeev Arora, and Chi Jin. 2025.
\newblock \href {https://arxiv.org/abs/2502.07640} {Goedel-prover: A frontier model for open-source automated theorem proving}.
\newblock \emph{Preprint}, arXiv:2502.07640.

\bibitem[{Megill and Wheeler(2019)}]{megill2019metamath}
Norman Megill and David~A Wheeler. 2019.
\newblock \emph{Metamath: a computer language for mathematical proofs}.
\newblock Lulu. com.

\bibitem[{Moura and Ullrich(2021)}]{moura2021lean}
Leonardo~de Moura and Sebastian Ullrich. 2021.
\newblock The lean 4 theorem prover and programming language.
\newblock In \emph{Automated Deduction--CADE 28: 28th International Conference on Automated Deduction, Virtual Event, July 12--15, 2021, Proceedings 28}, pages 625--635. Springer.

\bibitem[{Newell and Simon(1956)}]{newell1956logic}
Allen Newell and Herbert Simon. 1956.
\newblock The logic theory machine--a complex information processing system.
\newblock \emph{IRE Transactions on information theory}, 2(3):61--79.

\bibitem[{{OpenAI}(2024)}]{openai2024reasoning}
{OpenAI}. 2024.
\newblock Learning to reason with llms.
\newblock \url{https://openai.com/index/learning-to-reason-with-llms/}.
\newblock Accessed: 2024-11-24.

\bibitem[{Paster et~al.(2023)Paster, Santos, Azerbayev, and Ba}]{paster2023openwebmath}
Keiran Paster, Marco~Dos Santos, Zhangir Azerbayev, and Jimmy Ba. 2023.
\newblock Openwebmath: An open dataset of high-quality mathematical web text.
\newblock \emph{arXiv preprint arXiv:2310.06786}.

\bibitem[{Paulson(1994)}]{paulson1994isabelle}
Lawrence~C Paulson. 1994.
\newblock \emph{Isabelle: A generic theorem prover}.
\newblock Springer.

\bibitem[{Rafailov et~al.(2023)Rafailov, Sharma, Mitchell, Manning, Ermon, and Finn}]{rafailov2023direct}
Rafael Rafailov, Archit Sharma, Eric Mitchell, Christopher~D Manning, Stefano Ermon, and Chelsea Finn. 2023.
\newblock Direct preference optimization: Your language model is secretly a reward model.
\newblock \emph{Advances in Neural Information Processing Systems}, 36:53728--53741.

\bibitem[{Ren et~al.(2025)Ren, Shao, Song, Xin, Wang, Zhao, Zhang, Fu, Zhu, Yang et~al.}]{ren2025deepseek}
ZZ~Ren, Zhihong Shao, Junxiao Song, Huajian Xin, Haocheng Wang, Wanjia Zhao, Liyue Zhang, Zhe Fu, Qihao Zhu, Dejian Yang, and 1 others. 2025.
\newblock Deepseek-prover-v2: Advancing formal mathematical reasoning via reinforcement learning for subgoal decomposition.
\newblock \emph{arXiv preprint arXiv:2504.21801}.

\bibitem[{Renshaw(2025)}]{compfiles}
David Renshaw. 2025.
\newblock {Compfiles: Catalog Of Math Problems Formalized In Lean}.
\newblock \url{https://github.com/dwrensha/compfiles}.

\bibitem[{Roziere et~al.(2023)Roziere, Gehring, Gloeckle, Sootla, Gat, Tan, Adi, Liu, Sauvestre, Remez et~al.}]{roziere2023code}
Baptiste Roziere, Jonas Gehring, Fabian Gloeckle, Sten Sootla, Itai Gat, Xiaoqing~Ellen Tan, Yossi Adi, Jingyu Liu, Romain Sauvestre, Tal Remez, and 1 others. 2023.
\newblock Code llama: Open foundation models for code.
\newblock \emph{arXiv preprint arXiv:2308.12950}.

\bibitem[{Schulman et~al.(2017)Schulman, Wolski, Dhariwal, Radford, and Klimov}]{schulman2017proximal}
John Schulman, Filip Wolski, Prafulla Dhariwal, Alec Radford, and Oleg Klimov. 2017.
\newblock Proximal policy optimization algorithms.
\newblock \emph{arXiv preprint arXiv:1707.06347}.

\bibitem[{Shao et~al.(2024)Shao, Wang, Zhu, Xu, Song, Zhang, Li, Wu, and Guo}]{shao2024deepseekmath}
Zhihong Shao, Peiyi Wang, Qihao Zhu, Runxin Xu, Junxiao Song, Mingchuan Zhang, YK~Li, Yu~Wu, and Daya Guo. 2024.
\newblock Deepseekmath: Pushing the limits of mathematical reasoning in open language models.
\newblock \emph{arXiv preprint arXiv:2402.03300}.

\bibitem[{Team(2025)}]{qwen3}
Qwen Team. 2025.
\newblock \href {https://qwenlm.github.io/blog/qwen3/} {Qwen3}.

\bibitem[{Wang et~al.(2025{\natexlab{a}})Wang, Unsal, Lin, Baksys, Liu, Santos, Sung, Vinyes, Ying, Zhu et~al.}]{wang2025kimina}
Haiming Wang, Mert Unsal, Xiaohan Lin, Mantas Baksys, Junqi Liu, Marco~Dos Santos, Flood Sung, Marina Vinyes, Zhenzhe Ying, Zekai Zhu, and 1 others. 2025{\natexlab{a}}.
\newblock Kimina-prover preview: Towards large formal reasoning models with reinforcement learning.
\newblock \emph{arXiv preprint arXiv:2504.11354}.

\bibitem[{Wang et~al.(2025{\natexlab{b}})Wang, Pan, Li, Zhang, Jia, Diao, Pi, Hu, and Zhang}]{wang2025ma}
Ruida Wang, Rui Pan, Yuxin Li, Jipeng Zhang, Yizhen Jia, Shizhe Diao, Renjie Pi, Junjie Hu, and Tong Zhang. 2025{\natexlab{b}}.
\newblock Ma-lot: Multi-agent lean-based long chain-of-thought reasoning enhances formal theorem proving.
\newblock \emph{arXiv preprint arXiv:2503.03205}.

\bibitem[{Wang et~al.(2024)Wang, Zhang, Jia, Pan, Diao, Pi, and Zhang}]{wang2024theoremllama}
Ruida Wang, Jipeng Zhang, Yizhen Jia, Rui Pan, Shizhe Diao, Renjie Pi, and Tong Zhang. 2024.
\newblock Theoremllama: Transforming general-purpose llms into lean4 experts.
\newblock \emph{arXiv preprint arXiv:2407.03203}.

\bibitem[{Wei et~al.(2022)Wei, Wang, Schuurmans, Bosma, Xia, Chi, Le, Zhou et~al.}]{wei2022chain}
Jason Wei, Xuezhi Wang, Dale Schuurmans, Maarten Bosma, Fei Xia, Ed~Chi, Quoc~V Le, Denny Zhou, and 1 others. 2022.
\newblock Chain-of-thought prompting elicits reasoning in large language models.
\newblock \emph{Advances in neural information processing systems}, 35:24824--24837.

\bibitem[{Wu et~al.(2024)Wu, Huang, Zhou, Ying, Wang, Lin, and Chen}]{wu2024internlm2}
Zijian Wu, Suozhi Huang, Zhejian Zhou, Huaiyuan Ying, Jiayu Wang, Dahua Lin, and Kai Chen. 2024.
\newblock Internlm2. 5-stepprover: Advancing automated theorem proving via expert iteration on large-scale lean problems.
\newblock \emph{arXiv preprint arXiv:2410.15700}.

\bibitem[{Xin et~al.(2024{\natexlab{a}})Xin, Guo, Shao, Ren, Zhu, Liu, Ruan, Li, and Liang}]{xin2024deepseek1}
Huajian Xin, Daya Guo, Zhihong Shao, Zhizhou Ren, Qihao Zhu, Bo~Liu, Chong Ruan, Wenda Li, and Xiaodan Liang. 2024{\natexlab{a}}.
\newblock Deepseek-prover: Advancing theorem proving in llms through large-scale synthetic data.
\newblock \emph{arXiv preprint arXiv:2405.14333}.

\bibitem[{Xin et~al.(2024{\natexlab{b}})Xin, Ren, Song, Shao, Zhao, Wang, Liu, Zhang, Lu, Du et~al.}]{xin2024deepseek}
Huajian Xin, ZZ~Ren, Junxiao Song, Zhihong Shao, Wanjia Zhao, Haocheng Wang, Bo~Liu, Liyue Zhang, Xuan Lu, Qiushi Du, and 1 others. 2024{\natexlab{b}}.
\newblock Deepseek-prover-v1. 5: Harnessing proof assistant feedback for reinforcement learning and monte-carlo tree search.
\newblock \emph{arXiv preprint arXiv:2408.08152}.

\bibitem[{Yang et~al.(2024)Yang, Swope, Gu, Chalamala, Song, Yu, Godil, Prenger, and Anandkumar}]{yang2024leandojo}
Kaiyu Yang, Aidan Swope, Alex Gu, Rahul Chalamala, Peiyang Song, Shixing Yu, Saad Godil, Ryan~J Prenger, and Animashree Anandkumar. 2024.
\newblock Leandojo: Theorem proving with retrieval-augmented language models.
\newblock \emph{Advances in Neural Information Processing Systems}, 36.

\bibitem[{Yao et~al.(2025)Yao, Wang, and Zhang}]{yao2025fans}
Jiarui Yao, Ruida Wang, and Tong Zhang. 2025.
\newblock Fans--formal answer selection for natural language math reasoning using lean4.
\newblock \emph{arXiv preprint arXiv:2503.03238}.

\bibitem[{Yue et~al.(2025)Yue, Chen, Lu, Zhao, Wang, Song, and Huang}]{yue2025does}
Yang Yue, Zhiqi Chen, Rui Lu, Andrew Zhao, Zhaokai Wang, Shiji Song, and Gao Huang. 2025.
\newblock Does reinforcement learning really incentivize reasoning capacity in llms beyond the base model?
\newblock \emph{arXiv preprint arXiv:2504.13837}.

\bibitem[{Zhang et~al.(2025)Zhang, Dong, Wu, Huang, Jia, Fernando, Shou, Zhang, and Liu}]{zhang2025physreason}
Xinyu Zhang, Yuxuan Dong, Yanrui Wu, Jiaxing Huang, Chengyou Jia, Basura Fernando, Mike~Zheng Shou, Lingling Zhang, and Jun Liu. 2025.
\newblock Physreason: A comprehensive benchmark towards physics-based reasoning.
\newblock \emph{arXiv preprint arXiv:2502.12054}.

\bibitem[{Zheng et~al.(2021)Zheng, Han, and Polu}]{zheng2021minif2f}
Kunhao Zheng, Jesse~Michael Han, and Stanislas Polu. 2021.
\newblock Minif2f: a cross-system benchmark for formal olympiad-level mathematics.
\newblock \emph{arXiv preprint arXiv:2109.00110}.

\bibitem[{Zhou et~al.(2024)Zhou, Staats, Li, Szegedy, Weinberger, and Wu}]{zhou2024don}
Jin~Peng Zhou, Charles Staats, Wenda Li, Christian Szegedy, Kilian~Q Weinberger, and Yuhuai Wu. 2024.
\newblock Don't trust: Verify--grounding llm quantitative reasoning with autoformalization.
\newblock \emph{arXiv preprint arXiv:2403.18120}.

\end{thebibliography}

\newpage
\appendix\label{sec:appendix}

\section{Discussion of AI usage}
This work used Copilot to assist code-writing and OpenAI's models to fix grammatical issues of the paper. All the ideas of the paper are original.

\section{Term chart}\label{appendix:term}

\begin{table*}[h]
    \centering
    \small
    \begin{tabular}{p{2cm} p{12cm}}
        \toprule
        \textbf{Term}                       & \textbf{Explaination} \\
        \midrule
        \raggedright\textbf{NL (Natural Language)}      & Refers to language that humans use in our daily life, often unable to perform auto step-by-step verification. The typical verification method for NL math is to check whether the answer is correct. \\
        \midrule
        \raggedright\textbf{FL (Formal Language)}       & A structured and mathematically precise representation of logic and proofs, which ensures rigorous verification and eliminates ambiguities present in NL reasoning. \\
        \midrule
        \raggedright\textbf{Lean4}                      & A functional programming language and interactive theorem prover developed for formalizing mathematics and verifying proofs. \\
        \midrule
        \raggedright\textbf{FL reasoner (or FL prover)} &  The model aims to write FL proofs with FL statements as input. Most provers also require a natural language statement as additional input to guide FL reasoning. Currently, Lean4 provers are fast advancing FL provers. In the sense of naming, in the context of FL works~\cite{wang2024theoremllama, ren2025deepseek, wu2024internlm2, wang2025kimina}, the FL reasoner is called a prover since its usage is to prove theorems of FL. In this paper, since we want to unify the FL reasoning and NL reasoning, we call the model FL reasoner in most cases. \\
        \midrule
        \raggedright\textbf{NL reasoner}                & The model aims to solve NL math problems with an NL question as input. The typical input for the reasoner is a question that requires a specific numerical or formulaic answer. \\
        \midrule
        \raggedright\textbf{QA problem in NL}           & The type of problem for most NL math training and evaluation datasets. The problem is an NL question that seeks a specific numerical or formulaic answer. \\
        \midrule
        \raggedright\textbf{FL problem (or FL statement)} & The type of problem for FL dataset. The problem is a set of conditions and goals that require formal proofs to complete the proof. \\
        \midrule
        \raggedright\textbf{Long CoT (Long Chain-of-Thought)} & The reasoning structure provided by OpenAI-o1~\cite{openai2024reasoning} and open-sourced by DeepSeek-R1~\cite{guo2025deepseek} that performs long and detailed thinking before making the final output. Different from traditional CoT, the Long CoT allows for multi-step logical reasoning before proof generation, reflection, and iterative refinement from self-checking results~\cite{wang2025ma}. \\
        \bottomrule
    \end{tabular}
    \caption{Term Chart}\label{tab:term}
\end{table*}

To help the reader better understand the terms, we provide a chart that explains every term, abbreviation, and corresponding tool in detail in Tab~\ref{tab:term}.

\section{Additional experiments}\label{appendix:add_exp}

We provide the results of additional experiments in this section.

\subsection{Problem format study}\label{add_exp:format}

In order to test that the improved reasoning capability of the \method framework comes from the unique FL reasoning capability of the FL reasoner, instead of the existence problems are relatively easy to solve. We perform an experiment of replacing the process of autoformalization and the FL prover with an NL solver to solve the existence problem in the \method. The pass@16 results are presented in Tab.~\ref{tab:format}. 

\textbf{\begin{table}[H]
    \centering
    \small
    {\begin{tabular}{ccc}
    \toprule
         \textbf{Method}                        & \textbf{AMC}              & \textbf{MATH-500} \\
         \midrule
         \textbf{NL baseline}                   & 79.52\%                   & 85.20\%   \\
         \textbf{solve NL existence problem}    & 53.01\%                   & 70.20\%  \\
         \textbf{\method}                       & \textbf{84.34\%}          & \textbf{89.80\%} \\
    \bottomrule
    \end{tabular}}
    \caption{Problem format study result, all results are pass@16}
    \label{tab:format}
\end{table}}

We can discover that simply changing the format of the problem into an existence-style proof problem does not improve the NL model performance. On the contrary, it significantly reduces the accuracy. Such a reduction is because the Qwen3 is not specially trained to answer proof questions. This indicates that the improvement observed with \method does not come from the problem reformatting alone, but from integration of the unique capability of the FL reasoner. 

\subsection{Additional Long CoT models}\label{add_exp:add_LongCoT}

To assess the generalizability of the \method framework across different NL reasoners, we conduct the experiment using DeepSeek-R1-Distilled-8B-0528~\cite{guo2025deepseek} as the general NL model in \method and compare it with pure NL reasoning. The results are presented in Tab.~\ref{tab:add_LongCoT}.

\textbf{\begin{table}[H]
    \centering
    \small
    \begin{tabular}{ccc}
    \toprule
         \textbf{Method}                        & \textbf{AMC}              & \textbf{MATH-500} \\
         \midrule
         \textbf{NL baseline}                   & 77.11\%                   & 86.20\%   \\
         \textbf{\method}                       & \textbf{81.93\%}          & \textbf{89.40\%} \\
         \textbf{Improvement}                   & 4.82\%                    & 3.20\% \\
    \bottomrule
    \end{tabular}
    \caption{Using DeepSeek-R1-Distilled-8B-0528 as NL model for NL baseline and \method under pass@16.}
    \label{tab:add_LongCoT}
\end{table}}

From the results, we can find that when using DeepSeek-R1 in the place of Qwen3, the \method framework still yields a consistent performance improvement over the NL baseline. This confirms that our framework can effectively integrate the unique capabilities of FL reasoners across different NL base models.

\subsection{Generalization of \method to other fields}\label{add_exp:generalize}

To validate the capability of the \method framework is not limited to the Math domain only, we extend the framework to the PhysReason-mini benchmark~\cite{zhang2025physreason}. It is a benchmark that consists of 200 high-school-level competition problems. We apply the first sub-question from each problem and replace the images with the provided captions to ensure that the NL baseline can process the inputs as pure tests. The pass@16 result for the NL baseline using Qwen3-8B is 22.50\% while the \method achieves \textbf{26.50\%}. The 4\% improvement over the NL baseline provides additional evidence for the huge potential of the generalizability of our approach.

\section{Details of models used}\label{appendix:detail_models}
The FL reasoner is a Lean4 theorem-proving model developed by the Numina and Kimi teams, introduced in~\citet{wang2025kimina}. The prover model was trained using large-scale Reinforcement Learning on Lean4 data to enable Long CoT reasoning of the model. This enables the model to firstly conduct a thorough analysis of the problem and then accurately construct a Lean4 proof based on the draft reasoning. It achieved a state-of-the-art performance across multiple theorem proving benchmarks. 

As for the autoformalizer, it was trained on problems from PutnamBench~\cite{gulati2024putnam}, MiniF2F~\cite{zheng2021minif2f}, ProofNet~\cite{azerbayev2023proofnet}, and Compfiles~\cite{compfiles}. \citet{wang2025kimina} fine-tunes a 7B base model to be an autoformalizer based on the collected training. Furthermore, they applied expert iteration based on Lean4 verification and LLM judgement using QwQ-32B. On a 1,000-problem human-curated test set, the autoformalizer achieves a one-shot accuracy of 66\%, and its Lean4 compilation success rate reaches over 90\%. 

\section{Discussion of verifiability of FL}\label{appendix:FL_verify}
In this work, we do not integrate the verifiability of FL as a direct mechanism for NL output validation, but as the foundation of the unique formal reasoning capability of the FL reasoner. The FL plays a crucial role in the training of FL reasoner through RL by ensuring step-by-step correctness, which is hard for NL reasoning. Our focus is on integrating such a unique capability of FL reasoner into NL math problem-solving, rather than directly using FL to validate or correct the NL-generated answers, which is left for future work to discuss.

\section{Motivation for randomly dropping FL results}\label{appendix:drop_FL}

The motivation for our design of randomly dropping some FL reasoners' results and directly using the NL reasoner to solve the problem is that some problems are inherently challenging for FL reasoners. This may come from the NL-FL alignment process or within formal reasoning itself. For example, in probability problems, it is often meaningless to prove the existence of a probability value. Similarly, in geometry problems, the required answer may be expressed in degrees, whereas Lean4 operates in radians, creating potential inconsistencies. Thus, to maintain the robustness and generality of the framework across diverse problem types, we randomly drop FL outputs in such scenarios, allowing the NL model to reason directly when formal logic is unsuitable.

Furthermore, since the autoformalizer in \method may fail to generate a valid Lean4 statement, we randomly drop some FL results and allow the LLM to directly answer the question to amend that problem. This design enables us to handle cases where NL-FL alignment is ill-posed or less meaningful.

\section{FL reasoner's performance given different NL and FL problems}\label{appendix:Mixed_input}
In this section, we demonstrate that the FL prover handles its NL and FL input separately in its Long CoT process during thinking. The input-output example is as follows. 

\begin{minted}[breaklines=True, bgcolor=lightgray, fontsize=\scriptsize]{md}
== Input ==
<|im_start|>system
You are an expert in mathematics and Lean 4.<|im_end|>
<|im_start|>user
Think about and solve the following problem step by step in Lean 4.
# Problem:Show that there are no integers $x$ and $y$ such that $4x^3 - 7y^3 = 2003$.
# Formal statement:
```lean4
import Mathlib
import Aesop

set_option maxHeartbeats 0

open BigOperators Real Nat Topology Rat

/-- Show that there are no integers $x$ and $y$ such that $4x^3 - 7y^3 = 2003$.-/
theorem mathd_algebra_478
  (b h v : \R)
  (h_0 : 0 < b \and 0 < h \and 0 < v)
  (h_1 : v = 1 / 3 * (b * h))
  (h_2 : b = 30)
  (h_3 : h = 13 / 2) :
  v = 65 :=
```
<|im_end|>
<|im_start|>assistant
== Output ==
<think>
# Solving the Equation $4x^3 - 7y^3 = 2003$ for Integers $x$ and $y$

To show that there are no integers $x$ and $y$ satisfying $4x^3 - 7y^3 = 2003$, we can approach this problem by considering the equation modulo certain integers to derive a contradiction.

First, let's think about how to approach this problem. We need to show that the equation $4x^3 - 7y^3 = 2003$ has no integer solutions. A common technique for solving Diophantine equations is to consider them modulo a small integer, as cubes modulo small numbers have a limited range of possible values.

Let's try modulo 7. When considering the equation modulo 7, we have:
- $4x^3 - 7y^3 \equiv 4x^3 \pmod{7}$
- $2003 \equiv 2003 \bmod 7$

Calculating $2003 \bmod 7$:
$2003 \div 7 = 286$ remainder $1$, so $2003 \equiv 1 \pmod{7}$.

Therefore, we need $4x^3 \equiv 1 \pmod{7}$, which simplifies to $x^3 \equiv 2 \pmod{7}$ (since $4^{-1} \equiv 2 \pmod{7}$ as $4 \cdot 2 = 8 \equiv 1 \pmod{7}$).

Now, let's check if $x^3 \equiv 2 \pmod{7}$ is possible. The possible values for $x^3 \pmod{7}$ can be computed by checking all residues modulo 7:
- $0^3 \equiv 0 \pmod{7}$
- $1^3 \equiv 1 \pmod{7}$
- $2^3 \equiv 8 \equiv 1 \pmod{7}$
- $3^3 \equiv 27 \equiv 6 \pmod{7}$
- $4^3 \equiv 64 \equiv 1 \pmod{7}$
- $5^3 \equiv 125 \equiv 6 \pmod{7}$
- $6^3 \equiv 216 \equiv 6 \pmod{7}$

So the possible values for $x^3 \pmod{7}$ are $\{0, 1, 6\}$. Since $2 \notin \{0, 1, 6\}$, it's impossible for $x^3 \equiv 2 \pmod{7}$. Therefore, there are no integer solutions to $4x^3 - 7y^3 = 2003$.

Let's formalize this in Lean 4:

```tactics
import Mathlib
import Aesop

set_option maxHeartbeats 0

open BigOperators Real Nat Topology Rat
```

We're importing the necessary libraries and opening namespaces for our proof. The `maxHeartbeats 0` option allows Lean to use as many computational resources as needed for this proof.

Now, let's define the theorem statement. The problem asks to show that there are no integers $x$ and $y$ such that $4x^3 - 7y^3 = 2003$. To formalize this, we can assume there exist real numbers $b$, $h$, and $v$ representing $x$, $y$, and the value of the expression, respectively, and derive a contradiction:

```tactics
/-- Show that there are no integers $x$ and $y$ such that $4x^3 - 7y^3 = 2003$.-/
theorem mathd_algebra_478
  (b h v : \R)
  (h_0 : 0 < b \and 0 < h \and 0 < v)
  (h_1 : v = 1 / 3 * (b * h))
  (h_2 : b = 30)
  (h_3 : h = 13 / 2) :
  v = 65 := by
```

We have variables $b$, $h$, and $v$ representing $x$, $y$, and $4x^3 - 7y^3$, respectively. The hypotheses give us that $b = 30$ and $h = 13/2$. Let's compute $v$ using these values:

From h_1: $v = \frac{1}{3} \cdot (b \cdot h)$
Substituting $b = 30$ and $h = 13/2$: 
$v = \frac{1}{3} \cdot (30 \cdot \frac{13}{2}) = \frac{1}{3} \cdot 75 = 25$

But we also have $v > 0$ from h_0, which is consistent with $v = 25$. However, we need to show that these values cannot satisfy $4x^3 - 7y^3 = 2003$.

Wait, there's a contradiction here. If $b = 30$ and $h = 13/2$, then $v = 25$, but $v$ should be $4x^3 - 7y^3 = 2003$. Let's check if $4 \cdot 30^3 - 7 \cdot (13/2)^3 = 2003$:

$4 \cdot 30^3 = 4 \cdot 27000 = 108000$
$7 \cdot (13/2)^3 = 7 \cdot \frac{2197}{8} = \frac{15379}{8} = 1922.375$
$108000 - 1922.375 = 106077.625 \neq 2003$

So these values cannot satisfy the equation. In fact, we can compute $v$ directly from h_1 with the given values:

```tactics
rw [h_2, h_3] at h_1
linarith
```

After substituting $b = 30$ and $h = 13/2$ into h_1, we get $v = \frac{1}{3} \cdot (30 \cdot \frac{13}{2}) = 25$. But from h_0, we have $v > 0$, which is consistent. However, $v$ should equal $2003$ if $4x^3 - 7y^3 = 2003$. 

The contradiction arises because the values given in h_2 and h_3 cannot satisfy the equation $4x^3 - 7y^3 = 2003$ when $v = 25$. 

Therefore, we have shown that assuming there exist integers $x$ and $y$ such that $4x^3 - 7y^3 = 2003$ leads to a contradiction. Hence, no such integers can exist.

The Lean proof is complete. We've shown that if $b = 30$ and $h = 13/2$, then $v = 25$, but $v$ must equal $2003$ for a solution to exist. Since $25 \neq 2003$, there is a contradiction. Therefore, no integers $x$ and $y$ satisfy $4x^3 - 7y^3 = 2003$.
</think>
```lean4
import Mathlib
import Aesop

set_option maxHeartbeats 0

open BigOperators Real Nat Topology Rat

/-- Show that there are no integers $x$ and $y$ such that $4x^3 - 7y^3 = 2003$.-/
theorem mathd_algebra_478
  (b h v : \R)
  (h_0 : 0 < b \and 0 < h \and 0 < v)
  (h_1 : v = 1 / 3 * (b * h))
  (h_2 : b = 30)
  (h_3 : h = 13 / 2) :
  v = 65 := by
    rw [h_2, h_3] at h_1
    linarith
```
\end{minted}

In the example, we can clearly observe that although the NL problem provided has no relation, the FL prover still solves both problems in its Long CoT process, indicating the robustness of such process of FL prover. It also serves as the foundation for our \textit{Mixed Problem Input} technique in Section~\ref{meth:FL}

\onecolumn

\section{Prompts used in generation}\label{appendix:prompt}

\subsection{QA-Existence problem translation prompt}\label{prompt:trans}

The following is an example of a few-shot prompt for the QA-Existence problem translation

\begin{minted}[breaklines, bgcolor=lightgray]{md}
<|im_start|>user
@ Question-answering problem:
```md
Convert the point $(0,3)$ in rectangular coordinates to polar coordinates. Enter your answer in the form $(r,\theta),$ where $r > 0$ and $0 \le \theta < 2 \pi.$
```

@ Existence problem:
```md
Prove the existence of the point in polar coordinates $(r,\theta),$ where $r > 0$ and $0 \le \theta < 2 \pi.$ for the point $(0,3)$ in rectangular coordinates.
```

===

@ Question-answering problem:
```md
Define
\[p = \sum_{k = 1}^\infty \\frac{1}{k^2} \quad \text{and} \quad q = \sum_{k = 1}^\infty \\frac{1}{k^3}.\]Find a way to write
\[\sum_{j = 1}^\infty \sum_{k = 1}^\infty \\frac{1}{(j + k)^3}\]in terms of $p$ and $q.$
```
@ Existence problem:
```md
Define
\[p = \sum_{k = 1}^\infty \\frac{1}{k^2} \quad \text{and} \quad q = \sum_{k = 1}^\infty \\frac{1}{k^3}.\] Prove that \[\sum_{j = 1}^\infty \sum_{k = 1}^\infty \\frac{1}{(j + k)^3}\] can be represented in terms of $p$ and $q$.
```
===
@ Question-answering problem:
```md
If $f(x) = \\frac{3x-2}{x-2}$, what is the value of $f(-2) +f(-1)+f(0)$? Express your answer as a common fraction.
```
@ Existence problem:
```md
Define $f(x) = \\frac{3x-2}{x-2}$. Prove that there exists $ y = f(-2) + f(-1) + f(0)$ and $y$ can be expressed as a common fraction.
```
===
@ Question-answering problem:
```md
<QA_problem_to_translate>
```

Based on the above examples, translate the QA problem into an existing problem, then put it in a markdown code block as in the examples.<|im_end|>
\end{minted}

\subsection{Autoformalization prompt}\label{prompt:auto}

The prompt we use for autoformalization from the NL existence problem to Lean4 statements is mostly similar to the prompt provided by Kimina-Autoformalizer~\cite{wang2025kimina}. The template is as follows

\begin{minted}[breaklines=True, bgcolor=lightgray]{md}
<|im_start|>system
You are an expert in mathematics and Lean 4.<|im_end|>
<|im_start|>user
Please combine the following theorems into a more advanced theorem. Use the following theorem names: <intended_name>
<NL_existance_problem><|im_end|>
\end{minted}

\subsection{Prover generation prompt}\label{prompt:prover}
The prompt we use to query the Lean4 prover with NL QA problem and FL existence problem to obtain the Long CoT reasoning with NL answers and FL proofs is as follows:
\begin{minted}[breaklines=True, bgcolor=lightgray]{md}
<|im_start|>system
You are an expert in mathematics and Lean 4.<|im_end|>
<|im_start|>user
Think about and solve the following problem step by step in Lean 4.
# Problem: <NL_QA_Problem>
# Formal statement:
```lean4
import Mathlib
import Aesop

set_option maxHeartbeats 0

open BigOperators Real Nat Topology Rat

/-- <NL_QA_problem> -/
<FL_existence_problem>
```<|im_end|>
\end{minted}

\subsection{Answer collection prompt}\label{prompt:ans}
The full prompt we use to query the general LLM to extract the answer to the NL question with FL Long CoT

\begin{minted}[breaklines=True, bgcolor=lightgray]{md}
<|im_start|>user
Find the answer to the following question in the provided long CoT content. Your answer should be in \boxed\{\} format.

Here is the question:
<NL_QA_problem>

The answer is contained in the following Long CoT content
<Long_CoT_from_prover>
<|im_end|>
\end{minted}

\section{Detailed Input-Output Examples}\label{appendix:in-iout}

\subsection{QA-Existence problem translation example}\label{in-out:trans}
\begin{minted}[breaklines, bgcolor=lightgray]{md}
== Input ==
<|im_start|>user
@ Question-answering problem:
```md
Convert the point $(0,3)$ in rectangular coordinates to polar coordinates. Enter your answer in the form $(r,\theta),$ where $r > 0$ and $0 \le \theta < 2 \pi.$
```

@ Existence problem:
```md
Prove the existence of the point in polar coordinates $(r,\theta),$ where $r > 0$ and $0 \le \theta < 2 \pi.$ for the point $(0,3)$ in rectangular coordinates.
```

===

@ Question-answering problem:
```md
Define
\[p = \sum_{k = 1}^\infty \frac{1}{k^2} \quad\text{and} \quad q = \sum_{k = 1}^\infty \frac{1}{k^3}.\]Find a way to write
\[\sum_{j = 1}^\infty \sum_{k = 1}^\infty \frac{1}{(j + k)^3}\]in terms of $p$ and $q.$
```

@ Existence problem:
```md
Define
\[p = \sum_{k = 1}^\infty \frac{1}{k^2} \quad\text{and} \quad q = \sum_{k = 1}^\infty \frac{1}{k^3}.\] Prove that \[\sum_{j = 1}^\infty \sum_{k = 1}^\infty \frac{1}{(j + k)^3}\] can be represented in terms of $p$ and $q$.
```

===

@ Question-answering problem:
```md
If $f(x) = \frac{3x-2}{x-2}$, what is the value of $f(-2) +f(-1)+f(0)$? Express your answer as a common fraction.
```

@ Existence problem:
```md
Define $f(x) = \frac{3x-2}{x-2}$. Prove that there exists $ y = f(-2) + f(-1) + f(0)$ and $y$ can be expressed as a common fraction.
```

===

@ Question-answering problem:
```md
If 4 daps = 7 yaps, and 5 yaps = 3 baps, how many daps equal 42 baps?
```

Based on the above examples, translate the QA problem into existence proble then, put it in a markdown code block as the examples.<|im_end|>
<|im_start|>assistant
== Output ==
<think>
Okay, let me try to figure out how to approach this. The user wants me to take each QA problem and convert it into an existence problem, following the examples provided. Let me look at the examples again to understand the pattern.

In the first example, the QA problem was converting rectangular coordinates to polar coordinates. The existence problem then was to prove that such a polar coordinate exists for the given rectangular point. So the existence problem is about proving that the conversion is possible under the given constraints (r>0 and 0 <= \theta < 2\pi).

Similarly, for the second example with the infinite series, the QA problem was to express a double sum in terms of p and q. The existence problem was to prove that such a representation is possible. So the existence problem is about proving that the expression can be written in terms of p and q, not just that it's a valid expression.

For the third example with the function f(x), the QA problem was to compute the sum of f at three points. The existence problem was to prove that y exists and can be expressed as a common fraction. So here, the existence problem is about ensuring that the sum exists and is a common fraction.

Now, applying this to the last QA problem: "If 4 daps = 7 yaps, and 5 yaps = 3 baps, how many daps equal 42 baps?" 

The QA problem is a unit conversion problem. The existence problem should then be to prove that there exists a number of daps equal to 42 baps, given the conversion rates. So the existence problem would need to show that such a conversion is possible, perhaps by demonstrating that the conversions are consistent and that the number can be calculated.

Let me structure this. The original QA problem is to find the number of daps equivalent to 42 baps. The existence problem would be to prove that such a number exists, given the conversion rates. So, the steps would involve showing that the conversion factors are consistent and that the calculation leads to a valid number.

Therefore, the existence problem would state that given the conversion rates between daps, yaps, and baps, there exists a number of daps equal to 42 baps, and that this number can be determined through the given relationships.

I need to make sure that the existence problem is phrased in a way that mirrors the examples. For instance, in the first example, the existence problem was about proving the existence of the polar coordinates. Here, it's about proving that the conversion exists and can be calculated.

So, the existence problem would be: "Define the conversion rates 4 daps = 7 yaps and 5 yaps = 3 baps. Prove that there exists a number of daps equal to 42 baps, and that this number can be determined through the given conversion rates."

But maybe more formally, as in the examples, it would be something like:

"Prove that there exists a number of daps equal to 42 baps, given that 4 daps = 7 yaps and 5 yaps = 3 baps."

Alternatively, to mirror the previous examples, maybe:

"Prove that the number of daps equal to 42 baps exists and can be determined using the given conversion rates."

But I need to check the exact structure of the examples. In the first existence problem, they mentioned "prove the existence of the point..." and the second one was "prove that the sum can be represented...". The third one was "prove that there exists y = ... and y can be expressed as a common fraction."

So the structure is: "Prove that [the desired quantity] exists and [some property about it]."

So for the unit conversion problem, the existence problem would be:

"Prove that there exists a number of daps equal to 42 baps, given that 4 daps = 7 yaps and 5 yaps = 3 baps, and that this number can be determined through the given conversion rates."

But to make it concise, maybe:

"Prove that there exists a number of daps equal to 42 baps, given that 4 daps = 7 yaps and 5 yaps = 3 baps."

Alternatively, perhaps the existence problem is to show that the conversion is possible, i.e., that the relationships are consistent and that the answer is a valid number.

So, putting it all together, the existence problem would be:

"Define the conversion rates 4 daps = 7 yaps and 5 yaps = 3 baps. Prove that there exists a number of daps equal to 42 baps, and that this number can be determined using the given conversion rates."

But I need to make sure that the existence problem is phrased in a way that mirrors the previous examples. Let me check the original examples again.

In the first existence problem, they said: "Prove the existence of the point in polar coordinates... for the point (0,3) in rectangular coordinates." So it's about existence of a specific point under certain conditions.

In the second, they said: "Prove that [the sum] can be represented in terms of p and q."

In the third: "Prove that there exists y = ... and y can be expressed as a common fraction."

So for the unit conversion problem, the existence problem would be: "Prove that there exists a number of daps equal to 42 baps, given the conversion rates 4 daps = 7 yaps and 5 yaps = 3 baps, and that this number can be determined through the given conversion rates."

But maybe even more concise: "Prove that there exists a number of daps equal to 42 baps, given that 4 daps = 7 yaps and 5 yaps = 3 baps."

I think that's acceptable. Now, translating this into markdown code block as per the examples.

So, the QA problem is:

"If 4 daps = 7 yaps, and 5 yaps = 3 baps, how many daps equal 42 baps?"

The existence problem would be:

"Prove that there exists a number of daps equal to 42 baps, given that 4 daps = 7 yaps and 5 yaps = 3 baps."

But maybe to mirror the previous examples more precisely, the existence problem might need to include the conversion rates in the definition. Let me check the previous examples.

In the first existence problem, the user defined the conversion (the rectangular coordinates) and then proved the existence. Similarly, in the second, they defined p and q, then proved the sum can be expressed in terms of them. In the third, they defined f(x), then proved that y exists and can be expressed as a common fraction.

So for the unit conversion problem, the existence problem should define the conversion rates first and then prove the existence.

Therefore, the existence problem would be:

"Define the conversion rates 4 daps = 7 yaps and 5 yaps = 3 baps. Prove that there exists a number of daps equal to 42 baps."

Alternatively, to match the structure of the previous examples more closely, perhaps:

"Define the conversion rates 4 daps = 7 yaps and 5 yaps = 3 baps. Prove that there exists a number of daps equal to 42 baps."

Yes, that seems to fit. So, the existence problem is about proving that such a number exists, given the conversion rates.

Therefore, the markdown code block would be:

```md
Define the conversion rates 4 daps = 7 yaps and 5 yaps = 3 baps. Prove that there exists a number of daps equal to 42 baps.
```

But to make sure that the existence problem is properly phrased as in the examples, maybe the user expects the existence problem to be phrased as "Prove the existence of [something]..." So perhaps:

"Prove the existence of a number of daps equal to 42 baps, given that 4 daps = 7 yaps and 5 yaps = 3 baps."

But I think the key is to have the existence problem mirror the structure of the previous examples. Since the QA problem is asking for a numerical answer, the existence problem would be about proving that such a number exists (i.e., that the conversion is possible and the answer is a real number, which in this case is an integer, but the existence is more about the possibility of the conversion rather than the specific value).

So, the final answer for the existence problem would be as above.
</think>

```md
Define the conversion rates 4 daps = 7 yaps and 5 yaps = 3 baps. Prove that there exists a number of daps equal to 42 baps.
```
\end{minted}

\subsection{Autoformalization example}\label{in-out:auto}

\begin{minted}[breaklines, bgcolor=lightgray]{md}
== Input ==
<|im_start|>system
You are an expert in mathematics and Lean 4.<|im_end|>
<|im_start|>user
Please combine the following theorems into a more advanced theorem. Use the following theorem names: math500_prealgebra_2086_1

Define the conversion rates 4 daps = 7 yaps and 5 yaps = 3 baps. Prove that there exists a number of daps equal to 42 baps, and that this number can be expressed as a common fraction.<|im_end|>
<|im_start|>assistant
== Output ==
```lean4
import Mathlib

theorem math500_prealgebra_2086_1
  (daps yaps baps : \R)
  (h_0 : 4 * daps = 7 * yaps)
  (h_1 : 5 * yaps = 3 * baps) :
  \exist x, 42 * baps = x * daps \and \exist q : \Q, x = q := by sorry
```
\end{minted}

\subsection{Prover generation example}\label{in-out:prover}

\begin{minted}[breaklines, bgcolor=lightgray]{md}
== Input ==
<|im_start|>system
You are an expert in mathematics and Lean 4.<|im_end|>
<|im_start|>user
Think about and solve the following problem step by step in Lean 4.
# Problem: If 4 daps = 7 yaps, and 5 yaps = 3 baps, how many daps equal 42 baps?
# Formal statement:
```lean4
import Mathlib
import Aesop

set_option maxHeartbeats 0

open BigOperators Real Nat Topology Rat

/-- If 4 daps = 7 yaps, and 5 yaps = 3 baps, how many daps equal 42 baps? -/
theorem math500_prealgebra_2086_1
  (daps yaps baps : \R)
  (h_0 : 4 * daps = 7 * yaps)
  (h_1 : 5 * yaps = 3 * baps) :
  \exist x, 42 * baps = x * daps \and \exist q : \Q, x = q := by sorry
```
<|im_end|>
<|im_start|>assistant
== Output ==
<think>
# Solving the Daps, Yaps, and Baps Problem in Lean 4

Let me tackle this algebra problem about converting between daps, yaps, and baps. We have:
- 4 daps = 7 yaps
- 5 yaps = 3 baps
And we need to find how many daps equal 42 baps.

First, I'll think about how to approach this. We need to convert baps to yaps, then yaps to daps.

From 5 yaps = 3 baps, we get that 1 yap = 3/5 bap. So to convert baps to yaps, we multiply by 5/3.
To convert yaps to daps, we use 4 daps = 7 yaps, which means 1 yap = 4/7 daps. So we multiply by 7/4.

So for 42 baps:
1. Convert to yaps: 42 × (5/3) = 70 yaps
2. Convert to daps: 70 × (4/7) = 40 daps

Now, let's translate this to Lean 4. We need to prove that 42 baps = 40 daps.

```tactics
import Mathlib
import Aesop

set_option maxHeartbeats 0

open BigOperators Real Nat Topology Rat
```

We start by setting up the problem with the given equations. We need to represent the conversion relationships:

```tactics
/-- If 4 daps = 7 yaps, and 5 yaps = 3 baps, how many daps equal 42 baps? -/
theorem math500_prealgebra_2086_1
  (daps yaps baps : \R)
  (h_0 : 4 * daps = 7 * yaps)
  (h_1 : 5 * yaps = 3 * baps) :
  \exist x, 42 * baps = x * daps \and \exist q : \Q, x = q := by
```

We need to find a value `x` such that `42 * baps = x * daps` and `x` is rational. Let's solve the equations to find `x`.

From h_0: 4 daps = 7 yaps, we get yaps = (4/7) daps
From h_1: 5 yaps = 3 baps, we get baps = (5/3) yaps

Substituting yaps from h_0 into h_1:
5 yaps = 3 baps
5 * (4/7 daps) = 3 baps
20/7 daps = 3 baps
1 bap = (20/7)/3 daps = 20/21 daps

So 42 baps = 42 * (20/21) daps = 40 daps

Let's prove this in Lean:

```tactics
use 40
constructor
```

We need to prove two things:
1. 42 * baps = 40 * daps
2. 40 is rational (trivial, since 40 is an integer)

Let's tackle the first part. From our equations:
4 daps = 7 yaps
5 yaps = 3 baps

We can solve for baps in terms of daps:
From h_0: yaps = (4/7) daps
From h_1: baps = (5/3) yaps = (5/3) * (4/7) daps = (20/21) daps

So 42 baps = 42 * (20/21) daps = 40 daps

In Lean, we need to manipulate the equations to show 42 * baps = 40 * daps:

```tactics
· -- We need to prove that 42 * baps = 40 * daps
  linarith
```

The Lean proof is remarkably concise. Let me explain why this works. The `linarith` tactic is powerful for solving linear arithmetic problems. It takes the given equations:
- h_0: 4 * daps = 7 * yaps
- h_1: 5 * yaps = 3 * baps

And it needs to prove that 42 * baps = 40 * daps.

Let's derive this manually to see how Lean is able to compute it:

From h_0: 4 daps = 7 yaps, we have yaps = (4/7) daps
From h_1: 5 yaps = 3 baps, substituting yaps gives 5 * (4/7 daps) = 3 baps
So 20/7 daps = 3 baps
Therefore, 1 bap = (20/7)/3 daps = 20/21 daps
So 42 baps = 42 * (20/21) daps = 40 daps

The equation 42 * baps = 40 * daps is exactly what Lean's `linarith` is proving. It's able to manipulate the original equations to derive this result directly. The key insight is that `linarith` can work with real numbers and solve systems of linear equations and inequalities. In this case, it's effectively solving for the relationship between baps and daps via yaps.

For the second part of the goal:

```tactics
· -- We need to prove that 40 is rational
  refine ⟨40, by norm_num⟩
```

This proves that 40 is indeed a rational number (which is obvious, but Lean requires us to explicitly provide a rational witness for the existential statement). The `norm_num` tactic proves that 40 = (40 : \Q), confirming 40 is rational.

In conclusion, we've shown that 42 baps = 40 daps, so the answer is 40. The Lean proof is elegant because it directly computes the answer using the given conversion equations, demonstrating how powerful automated reasoning can be for these types of algebra problems.
</think>
```lean4
import Mathlib
import Aesop

set_option maxHeartbeats 0

open BigOperators Real Nat Topology Rat

/-- If 4 daps = 7 yaps, and 5 yaps = 3 baps, how many daps equal 42 baps? -/
theorem math500_prealgebra_2086_1
  (daps yaps baps : \R)
  (h_0 : 4 * daps = 7 * yaps)
  (h_1 : 5 * yaps = 3 * baps) :
  \exist x, 42 * baps = x * daps \and \exist q : \Q, x = q := by
  use 40
  constructor
  · -- We need to prove that 42 * baps = 40 * daps
    linarith
  · -- We need to prove that 40 is rational
    refine ⟨40, by norm_num⟩
```
\end{minted}

\subsection{Answer collection example}\label{in-out:ans}

\begin{minted}[breaklines, bgcolor=lightgray]{md}
== Input ==
<|im_start|>user
Find the answer to the following question in the provided long CoT content. Your answer should be in \boxed\{\} format.

Here is the question:
If 4 daps = 7 yaps, and 5 yaps = 3 baps, how many daps equal 42 baps?

The answer is contained in the following Long CoT content
<Long_CoT_content_as_above>
<|im_end|>
<|im_start|>assistant
<think>

</think>
== Output ==

To solve the problem, we need to convert 42 baps into daps using the given conversion rates:

- 4 daps = 7 yaps → 1 yap = $\frac{4}{7}$ daps
- 5 yaps = 3 baps → 1 bap = $\frac{5}{3}$ yaps

We first convert baps to yaps:
$$
42 \text{ baps} \times \frac{5}{3} \text{ yaps/bap} = 70 \text{ yaps}
$$

Next, we convert yaps to daps:
$$
70 \text{ yaps} \times \frac{4}{7} \text{ daps/yap} = 40 \text{ daps}
$$

Thus, 42 baps equal 40 daps.

$$
\boxed{40}
$$
\end{minted}

\newpage

\end{document}